\documentclass{article}

    \usepackage[preprint]{neurips_2025}

\usepackage{amssymb, amsmath,stackengine}
\usepackage[export]{adjustbox}
\usepackage{cancel}

\usepackage[utf8]{inputenc} 
\usepackage[T1]{fontenc}    
\usepackage{hyperref}       
\usepackage{url}            
\usepackage{booktabs}       
\usepackage{amsfonts}       
\usepackage{nicefrac}       
\usepackage{microtype}      
\usepackage{xcolor}         
\usepackage{amsthm}
\usepackage{graphicx}
\usepackage{graphics}
\usepackage{multirow}
\usepackage{enumitem}

\usepackage{amsthm}
\usepackage{caption}

\theoremstyle{definition}
\newtheorem{definition}{Definition}

\theoremstyle{lemma}

\theoremstyle{proposition}

\definecolor{blind1_blue}{HTML}{005AB5}
\definecolor{blind1_red}{HTML}{DC3220}

\definecolor{blind2_blue}{HTML}{1A85FF}
\definecolor{blind2_violet}{HTML}{D41159}

\usepackage[utf8]{inputenc}
\usepackage[T1]{fontenc}
\usepackage{lipsum}
\usepackage{algorithm}
\usepackage[noend]{algpseudocode}
\usepackage[compatibility=false]{caption}
\DeclareCaptionFormat{ruled}{\leavevmode\leaders\hrule height 0.8pt depth0pt\hfill\mbox{}\endgraf#1#2 #3 \vspace{-0.4\baselineskip}\leavevmode\leaders\hrule height 0.6pt\hfill\null\vspace*{-0.6\baselineskip}}
 \algrenewcommand\algorithmiccomment[1]{\hfill\({}\triangleright{}\){\footnotesize#1}}%

\algrenewtext{For}[1]{\textbf{for} #1}
\algrenewtext{While}[1]{\textbf{while} #1}
\algrenewtext{ForAll}[1]{\textbf{for all} #1}
\algrenewtext{If}[1]{\textbf{if} #1}
\algrenewtext{ElseIf}[1]{\textbf{else if} #1}
\algrenewtext{Else}{\textbf{else}}

\usepackage{tablefootnote}

\title{Wasserstein Hypergraph Neural Network}

\author{Iulia Duta \\
  Department of Computer Science\\
  University of Cambridge\\
  \texttt{id366@cam.ac.uk} \\
  \And
  Pietro Li\`{o} \\
  Department of Computer Science\\
  University of Cambridge\\
  \texttt{pl219@cam.ac.uk} \\
}

\begin{document}

\maketitle

\begin{abstract}
The ability to model relational information using machine learning has driven advancements across various domains, from medicine to social science. While graph representation learning has become mainstream over the past decade, representing higher-order relationships through hypergraphs is rapidly gaining momentum.
In the last few years, numerous hypergraph neural networks have emerged, most of them falling under a two-stage, set-based framework. The messages are sent from nodes to edges and then from edges to nodes. However, most of the advancement still takes inspiration from the graph counterpart, often simplifying the aggregations to basic pooling operations.
In this paper we are introducing Wasserstein Hypergraph Neural Network, a model that treats the nodes and hyperedge neighbourhood as distributions and aggregate the information using Sliced Wasserstein Pooling. Unlike conventional aggregators such as mean or sum, which only capture first-order statistics, our approach has the ability to preserve geometric properties like the shape and spread of distributions. This enables the learned embeddings to reflect how easily one hyperedge distribution can be transformed into another, following principles of optimal transport.
Experimental results demonstrate that applying Wasserstein pooling in a hypergraph setting significantly benefits node classification tasks, achieving top performance on several real-world datasets.

%

\end{abstract}

\section{Introduction}


The potential to learn from relational data has substantially broadened the applicability of machine learning, extending its reach to a wide range of fields from medicine~\cite{journals/corr/abs-2310-13767,pred_pat_outcome} , to physics~\cite{gonzalez20a_battaglia_simulate,lam2023graphcastlearningskillfulmediumrange}, social science~\cite{monti2019fake_news} and chemistry~\cite{pmlr-v70-gilmer17a, huang2020skipgnn_molecular_interactions}. The flexibility of graph structures makes them well-suited for representing complex natural phenomena involving various types of interactions. As a result, graphs quickly became synonymous with modelling interactions. However, while graphs are restricted to model pairwise connections, many real-world interactions involve more than two entities. To fill this gap, a generalisation of graphs called hypergraphs were introduced, allowing for the representation of higher-order relationships among multiple elements.


More precisely, a hypergraph is characterized by a set of edges, where each edge connects a set of nodes, potentially of varying cardinality.  The challenge of designing hypergraph neural networks becomes the challenge of properly modelling these sets. Many approaches~\cite{allset, wang2022equivariant, UniGNN} tackle this using a two-step process: first, the model aggregates information from the nodes within each hyperedge to compute a representation for that hyperedge. Then, in the second step, it updates each node’s representation using information from the hyperedges it belongs to. Both steps rely on methods designed to handle sets of elements.

Although set representation learning has seen significant progress in recent years~\cite{set_survey}, hypergraph networks still largely rely on sum-based aggregation methods such as Deep Sets~\cite{DeepSets} and Set Transformers~\cite{set_transformer}. Despite their strong theoretical foundation, these methods can struggle to effectively capture the full geometry of set-structured inputs~\cite{Wass_pooling}.


In this work we are introducing Wasserstein Hypergraph Neural Networks (WHNN), a class of hypergraph models that uses Sliced Wasserstein Pooling (SWP) ~\citep{Wass_pooling} as a node and hyperedge aggregator. This pooling is based on the Wasserstein distance - an optimal transport metric which measures the distance between two distributions based on the cost of transporting mass from one to another. SWP treats the set elements as samples from an underlying distribution and generates a vector representation that captures geometric relationships between inputs, such as shape, spread, and density. This information is often lost when aggregating using summation. 

We argue that this geometric information is highly relevant for hypergraph learning. Our experimental results support this claim, showing that WHNN not only outperforms traditional set-based aggregation methods used in previous hypergraph models, but also achieves superior performance compared to several strong hypergraph methods across a range of real-world datasets.

\textbf{Our main contributions} are summarised as follow: 

\begin{enumerate}

\item We propose \textbf{a novel hypergraph architecture, the Wasserstein Hypergraph Neural Network (WHNN)}, which leverages Sliced Wasserstein Pooling for both node and hyperedge aggregation to more effectively capture the geometric structure of the feature space.

\item We empirically show that Wasserstein aggregation is highly effective for hypergraph representation, consistently \textbf{outperforming traditional sum-based methods} such as Deep Sets~\cite{DeepSets} and Set Transformers~\cite{set_transformer}, regardless of the encoder used to process the nodes.

\item Wasserstein Hypergraph Neural Network achieves top results on multiple real-world datasets, highlighting the advantages of incorporating optimal transport into hypergraph processing.
\end{enumerate}

The paper is structured as follows: Sections~\ref{related_work} and ~\ref{background} describe similar efforts and introduce background concepts. Sections~\ref{model} introduces the architectures and innovative aspects of the methodology. Then, Section~\ref{experiments} contains and discusses experiments supporting the claimed contributions.






\section{Related Work}
\label{related_work}

\paragraph{Hypergraph representation learning.}
    Hypergraphs represent a versatile structure for modeling group-wise interactions, which allows us to capture interactions between various number of elements. This flexibility, combined with the widespread presence of higher-order interactions in real-world scenarios, has led to a growing interest in developing machine learning architectures for modeling hypergraph data. Some methods~\cite{HyperNN,SHNN} reduce the hypergraph to a clique-expansion graph that can be further processed with standard graph neural networks. A more popular approach is based on a two-stage framework~\cite{allset, UniGNN}, which sends the information from node to hyperedges and then from hyperedges back to nodes. Depending on how these stages are instantiated, several architectures emerged. HCHA and HERALD~\citep{HCHA, HERALD} use an  attention mechanism to combine the information,  AllDeepSets~\citep{allset} uses Deep Set model, while AllSetTransformer~\citep{allset} is using a PMA-like~\cite{set_transformer} pooling. 
    
    In all of these methods the information sent from the node is independent of the target hyperedge. Recently, models that create edge-dependent node representation have gained traction. ED-HNN~\cite{wang2022equivariant} uses as messages a concatenation of node and hyperedge information, while MultiSetMixer~\cite{telyatnikov2025hypergraph} uses MLP-Mixer~\cite{tolstikhin2021mlpmixer} to combine the information. Similar to our node encoder, CoNHD\cite{zheng2025corepresentation} incorporates pairwise propagation at the hyperedge-level using self-attention blocks (SAB~\cite{set_transformer}) to create edge-dependent representations. However, similar to~\cite{whatsnet}, the model is only tested on hyperedge-dependent node classification tasks, where each node is assigned multiple labels corresponding to the number of hyperedges it participates in. A complementary line of work~\cite{thnn} is representing uniform hypergraphs as high-dimensional tensors and applies tensorial operators to propagate the information.  
    
    In contrast, we are interpreting the hyperedges as samples from a set of probability distributions, and uses Sliced Wasserstein Pooling to aggregate the information such that we preserve geometric information. In terms of node encoders, we are experimenting with both edge-dependent and edge-independent modules.

\paragraph{Set representation learning.} The core operation in set representation learning is the permutation-invariant operator that aggregate the information without imposing an order among elements. Popular examples of such operator include summation, mean or maximum. More recently, learnable version of permutation-invariant poolings were introduced. Among these, Deep Sets~\cite{DeepSets} is using element-wise encoding of the elements followed by summation and is proved to be universal approximator for permutation-invariant functions. Janossy Pooling~\cite{murphy2018janossy} extends this model by explicitly aggregating pairs of elements. On the other hand, Set Transformer~\cite{set_transformer} and~\cite{Skianis2019RepTS} uses an anchor set as a reference and compute the similarity against this set as a representation, while FSPool~\cite{Zhang2019FSPool} sorts the elements feature-wise to create a canonical order. Recently, ~\cite{equiv_vs_invariant} shows empirically that combining an equivariant backbone with an invariant pooling layer creates powerful set representation learning. Inspired by optimal transport literature, Sliced Wasserstein Pooling was introduced in ~\cite{Wass_pooling} as a geometrically-interpretable set representation technique. 

    
\paragraph{Wasserstein embeddings.}
In recent years, Wasserstein distance has attracted significant attention in deep learning, demonstrating success in areas such as generative modeling~\cite{WGAN, nguyen2021distributional}, natural language processing~\cite{frogner2019learningembeddingsentropicwasserstein} and point cloud processing~\cite{point_cloud_Wass}. In graph representation learning, Wasserstein distance was used to define a similarity kernel between pair of graphs~\cite{Togninalli19}. While recognised as a powerful tool, computing this distance for each pair of compared graphs is extremely inefficient. More recent works~\cite{WGRL,OTK,courty2018learning} try to reduce this cost by introducing Wasserstein embeddings. The purpose of a Wasserstein embedding is to infer a vector representation such that the $L_2$ distance in the vector space approximates the Wasserstein distance in the input space. Particularly important for us is the work of ~\cite{Wass_pooling} which produces set representations using efficient Wasserstein embeddings. 

To more effectively capture the internal structure of node and hyperedge neighborhoods, we employ Sliced Wasserstein Pooling as the aggregation operator in hypergraph message passing, demonstrating its advantages for hypergraph representation learning.



    


    

\section{Background}
\label{background}

\subsection{Hypergraph Representation Learning}
\label{framework}
    A hypergraph is a tuple $\mathcal H = (V, E)$ where $V=\{v_1, v_2 \dots v_N\}$ is a set of nodes, and $E=\{e_1, e_2 \dots e_M\}$ is a set of hyperedges. Different than the graph structure, where each edge contains exactly two nodes, in a hypergraph each hyperedge contains a set of nodes which can vary in cardinality. Each node  $v_i$ is characterize by a feature vector $x_i \in \mathbb{R}^d$. We denote by \textit{neighbourhood of hyperedge $e_i$} the set of nodes that are part of that hyperedge $\{v_j | v_j \in e_i \}$. Similarly, the \textit{neighbourhood of a node $v_i$} is the set of all hyperedges containing that node $\mathcal{N}_{v_i}=\{e_j | v_i \in e_j\}$.

    Several architectures were developed for hypergraph-structured input~\cite{HyperNN,wang2022equivariant,ijcai21-UniGNN,allset}. However, the most general pipeline follow a two-stage framework, inspired by the bipartite representation of the hypergraphs.
    First, the information is sent from nodes to the hyperedges using a permutation-invariant operator $z_j = f_{V\rightarrow E}(\{x_i|v_i \in e_j\})$. Secondly, the messages are sent back from hyperedge to nodes $\tilde{x}_i = f_{E\rightarrow V}(\{z_j| v_i \in e_j \})$. 

    %

\begin{table}[t]
\centering
\caption{The update rules used as aggregation steps in various hypergraph neural networks from the literature. As shown by the equations, all of them relies either on (weighted) summation or mean to combine the information. While theoretically powerful, summing can easily destroy all the geometric relationships between points. $\mathcal{N}_v(i)$ is the neighbourhood of node $v_i$ of cardinality $d_i$,  $\mathcal{N}_e(j)$ is the neighbourhood of edge $e_j$ of cardinality $d_j$ and $\epsilon$, $W_*$, $\tilde{W}_*$ are learnable parameters.} 
\vspace{1.5mm}
\label{tab:updates}
\renewcommand{\arraystretch}{1.6}
\begin{tabular}{clllllllll}
\hline
\textbf{Model} & \textbf{Hyperedge aggregation} & \textbf{Node aggregation} \\
\hline
HGNN & 
$h_e \gets \sum_{i \in \mathcal{N}_e(e)} \frac{1}{\sqrt{d_i}} x_iW$ & 
$x_i \gets \frac{1}{\sqrt{d_i}} \sum_{e \in \mathcal{N}_v(i)} \frac{1}{d_e} h_e$ \\
HCHA\tablefootnote{The coefficients $\alpha_{e,i}$ used in summations are scalars predicted as $MLP(x_i||h_e)$} & 
$h_e \gets \sum_{i \in \mathcal{N}_e(e)} \alpha_{e,i} x_iW$ & 
$x_i \gets \sum_{e \in \mathcal{N}_v(i)} \tilde{\alpha}_{i,e} h_e \tilde{W}$ \\
UniGIN & 
$h_e \gets \sum_{i \in \mathcal{N}_e(e)} x_i$ & 
$x_i \gets \sum_{e \in \mathcal{N}_v(i)} h_eW + (1+\epsilon)x_iW$ \\
ED-HNN & 
$h_e \gets \sum_{i \in \mathcal{N}_e(e)} \text{MLP}(x_u)$ &  
$x_i \gets \sum_{e \in \mathcal{N}_v(i)} \text{MLP}(x_i \| h_e)$ \\
AllDeepSets & 
$h_e \gets \text{MLP}(\sum_{i \in \mathcal{N}_e(e)} \text{MLP}(x_u))$ & 
$x_i \gets \text{MLP}(\sum_{e \in \mathcal{N}_v(i)} \text{MLP}(h_e))$ \\
AllSetTransformer\tablefootnote{The function $\sigma$ is a combination of residual connections and layer normalisations, while  $\alpha_{i}=(\theta W_q)(x_i W_k)^T $ with $\theta$, $W_q$ and $W_k$ as learnable parameters. } & 
$h_e \gets \sigma(\sum_{i \in \mathcal{N}_e(e)}(\alpha_i x_iW_v)$ & 
$x_i \gets \sigma(\sum_{e \in \mathcal{N}_v(i)}(\tilde{\alpha}_e h_e\tilde{W}_v)$ \\
\hline
\end{tabular}
\end{table}

    While aggregators like Deep Sets~\cite{DeepSets} were theoretically capable of approximating any permutation-invariant function on sets, it relies on the initial encoder (such as MLPs) to reshape the feature space in a way in which the sum pooling is not losing important information. In other words, it moves the complexity of the representation from the pooling to the initial encoding. This is in line with the empirical results shown in ~\cite{equiv_vs_invariant} where, in order to preserve good performance,  mean pooling requires more complex encoders compared to more sophisticated pooling methods.


    
    In this work we are following the standard two-stage framework. Compared to existing methods, we take advantage of the success demonstrated by Sliced Wasserstein Pooling in capturing and retaining the geometric structure of sets and proposed the first hypergraph model that uses optimal transport techniques to perform the node and hyperedge aggregation. 

\subsection{Sliced Wasserstein Pooling (SWP)}
    \label{sec:Wass}
    
    To ensure the method's readability, this section introduces all the key concepts underlying our Wasserstein Hypergraph Neural Network. 
    First, we will define the 2-Wasserstein metric, approximate it using the tractable Sliced-Wasserstein distance and finally present the algorithm to compute the SWP used as an aggregator in our model.
    
    \begin{definition}
        \vspace{1mm}
        The \textbf{2-Wasserstein distance} between two distributions $p_i$ and $p_j$ over $\mathbb{R}^d$ is defined as:
        \begin{equation}
        \mathcal{W}_2(p_i, p_j) = \Big( \inf_{\gamma \in \Gamma (p_i, p_j)} \int_{\mathbb{R}^n \times \mathbb{R}^n} ||x - y ||^2 d \gamma (x,y) \Big)^{\frac{1}{2}},
        \end{equation}
        where $ \Gamma (p_i, p_j)$ represent the collection of all the transport plans with marginals $p_i$ and $p_j$.
    \end{definition}

    In simpler terms, the 2-Wasserstein distance quantifies the cost of transforming one distribution into another.
    
    Unfortunately, computing the infimum over all possible transport maps is generally untractable. However, \textbf{in the one dimensional case (when $d=1$) ), a closed-form solution exists} that avoids expensive optimization. Specifically, when $p_i$ and $p_j$ are probability distributions over $\mathbb{R}$, the 2-Wasserstein distance is given by $\mathcal{W}_2(p_i, p_j) = \Big (\int_0^1 | F_{p_i}^{-1}(t) - F_{p_j}^{-1}(t)|^2 dt \Big) ^{\frac{1}{2}}$, where $F_{p_i}^{-1}$ and $F_{p_j}^{-1}$ denote the inverse cumulative distribution functions of $p_i$ and $p_j$. A key practical benefit of this formulation is that this inner integral can be empirically estimated using a discrete sum over sorted samples from the distribution.

    Building on this observation, Sliced Wasserstein distance~\cite{sliced_wass} was introduce to approximate the Wasserstein distance, by projecting the high-dimensional probabilities into 1D lines using all possible directions on the unit sphere.
    
\begin{definition}
    \vspace{1mm}
    The \textbf{Sliced Wasserstein distance} between two distributions $p_i$ and $p_j$ over $\mathbb{R}^d$ is defined as:
        \begin{equation}
        \mathcal{SW}_2(p_i, p_j) = \Big( \int_{S^{d-1}} \mathcal{W}_2(P_{\theta} p_i, P_{\theta} p_j)d \theta \Big)^{\frac{1}{2}} \approx \Big (\frac{1}{L} \sum_{l=1}^L \underbrace{{\mathcal{W}_2(P_{\theta_l} p_i, P_{\theta_l} p_j)}}_{\text{\textcolor{gray}{1D Wasserstein distance}}} \Big)^{\frac{1}{2}},
        \end{equation}
    where $S^{d-1}$ is the unit sphere in $\mathbb{R}^d$,  $P_{\theta} p_i$ represent the projection (pushforward) of $p_i$ onto the line direction $\theta$ and $\{ \theta_l \}_{l=1}^L$ represents the set of $L$ directions used to empirically approximate the expectation. 
\end{definition}

To avoid the computational cost of calculating distances between every pair of probability distributions, the \textbf{Sliced Wasserstein embedding}~\cite{Wass_pooling} was introduced. It maps a probability distributions $p_i$ to a vector $\phi(p_i)$ in such a way that the Euclidean distance between the vectors (which is inexpensive to compute)  approximates the Sliced Wasserstein distance between the original distributions $||\phi(p_i)-\phi(p_j)||_2 \approx \mathcal{SW}_2(p_i,p_j)$. In other words, it provides a vectorial representation that captures the geometric structure of distributions, preserving information about how costly it is to transform one distribution into another. This geometric encoding reflects characteristics such as shape, spread, and density. This proves useful in our context, as it allows us to quantify the cost of transforming one hyperedge into another — a measure we argue effectively captures the similarity between group interactions (hyperedges).

Since our nodes and hyperedges are sets rather than distributions, we use a variant of this embedding called \textbf{Sliced Wasserstein Pooling}~\cite{Wass_pooling}, which is designed not as an embedding of probability distributions themselves, but rather as an embedding of sets sampled from those distributions.
In short, Sliced Wasserstein Pooling encodes a set of points by measuring, in an efficient way, how different they are positioned compared to a set of reference points. The complete algorithm as used in our model is described in the following section.










\begin{figure}[t]
    \centering
    \vspace{-3mm}
     \includegraphics[width=1.0\textwidth]{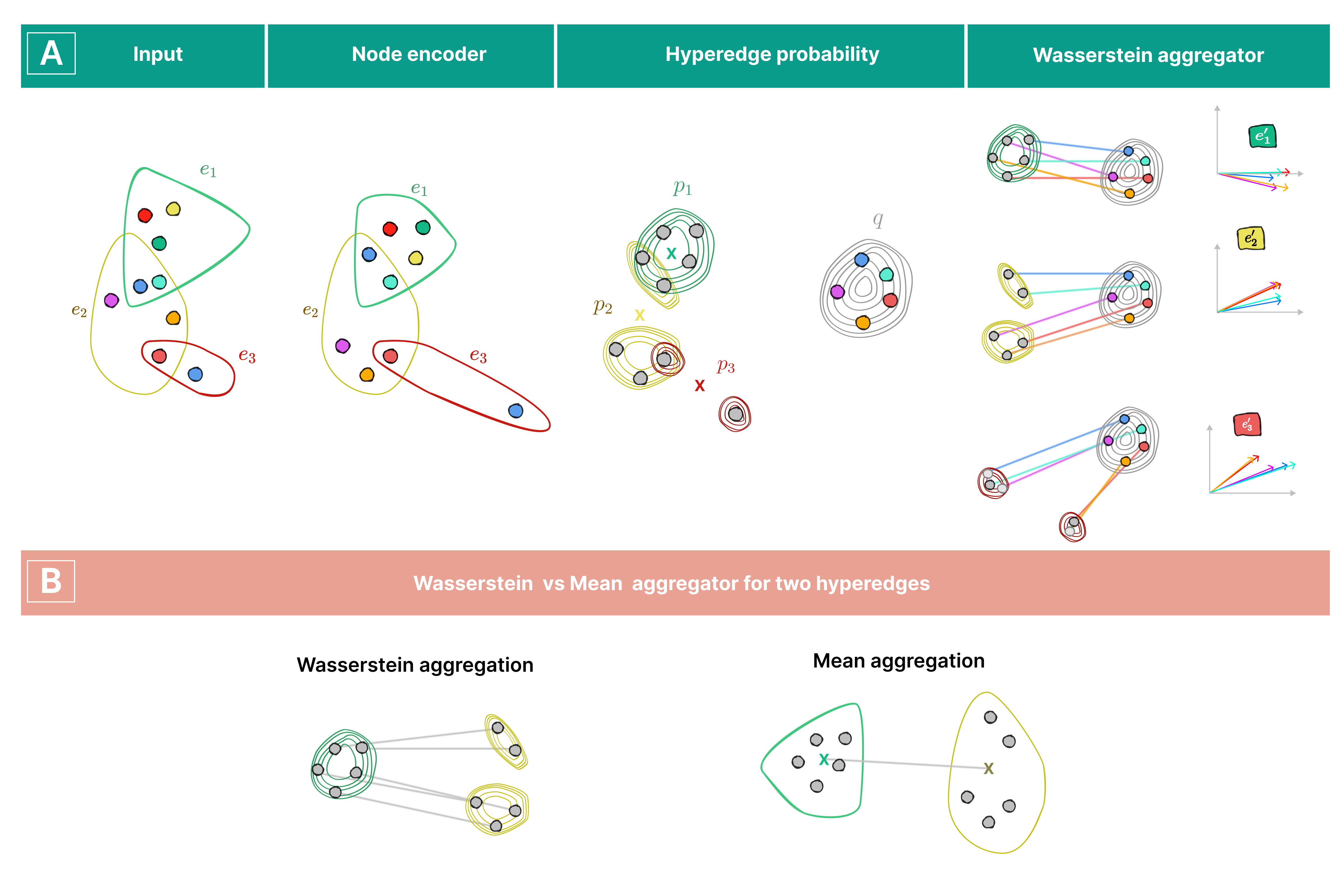}
     
     \caption{\textbf{(A) One stage (node-to-hyperedge) of Wasserstein Hypergraph Neural Network pipeline} designed to be more sensitive to the geometric structure of the hyperedge compared to the traditional aggregators. First, a \textbf{node encoder} processes the nodes using a simple MLP or an edge-dependent self-attention block (SAB). The hypergraphs is than viewed as a \textbf{collection of probability distributions} $\{ p_i \}$, one for each hyperedge, with the observed nodes treated as samples drawn from it. An additional distribution $q$ is picked as a reference. Finally the \textbf{Sliced Wasserstein Pooling} is adopted as an aggregation method: each hyperedge is represented by its Sliced Wasserstein distance to a reference distribution. \textbf{(B)} Compared to the standard mean pooling which only capture the difference between the mean of the distributions (visualised as a cross), the euclidean distance between the obtained hyperedges quantify the cost of transforming one group into another.} 
     \label{fig:main}
    \vspace{-4mm}
\end{figure}
\section{Wasserstein Hypergraph Neural Network}
\label{model}

Taking inspiration from the success of Wasserstein embeddings in set representation learning ~\cite{Wass_pooling, OTK}, 
we are introducing Wasserstein Hypergraph Neural Network (WHNN), a neural network for processing hypergraph structured data which replace the standard (weighted) mean aggregator with SWP, thus better capturing the internal structure of the neighbourhoods.

The model follows the two-stage framework introduced in Section~\ref{framework}, by sending information from nodes to hyperedges and vice-versa. For simplicity this section only describes the nodes to hyperedges mechanism, as the hyperedge-to-node operation is entirely symmetrical. The entire pipeline is depicted in Figure~\ref{fig:main} and Algorithm~\ref{main_algorithm}. For readability, the algorithm is presented sequentially for each hyperedge. However our implementation processes all hyperedges in parallel. 

First we will project the node features into a more expressive representation. Each hyperedge is then associated with a probability distribution, with its constituent nodes treated as samples. These distributions are embedded using a Wasserstein-based aggregator to obtain the final hyperedge representations. These hyperedge representations are fed into the hyperedges-to-nodes stage. 


\paragraph{Node encoder.} The goal of this module is to enhance the representation of node features by projecting them into a more informative space. We are experimenting with two types of encoders: an \textit{edge-independent} one where the node is carrying the same representation in each hyperedge it is contained, and an \textit{edge-dependent} one which takes into account pairwise interactions.

The edge-independent encoder is a simple MLP, which is applied in parallel for each node. This way a node $i$ is characterized by the same feature vector in each hyperedge $e$ it is part of.
\begin{equation*}
    \tilde{x}_i^e = MLP(x_i)
\end{equation*}
On the other hand, for the edge-dependent encoder, each node has a different representation in each hyperedge it is part of. To achieve this, for each hyperedge, we are using a Set Attention Block layer (SAB) as introduce in~\cite{set_transformer} which propagates the information between each pair of two nodes contained in that hyperedge. The full version of the block acts as follow:
\begin{align*}
    z_i^e &= \sigma(x_i + \sum_{j \in e}(x_iW_q)(x_jW_k)^T(x_jW_v)) \\
    \tilde{x}_i^e &= \sigma(z_i^e + MLP(z_i^e)),
\end{align*}
where $\sigma$ denote layer normalisation and $W_k$, $W_q$ and $W_v \in \mathbb{R}^{d \times d}$ are learnable parameters.

\begin{figure}[t]
{
\captionsetup{format=ruled, labelfont=sc}
   \begin{minipage}[t]{.47\textwidth}
    \captionof{algorithm}{One Layer of Wasserstein Hypergraph Neural Network$^*$}
    \label{main_algorithm}
    \begin{algorithmic}[1]
      \State \textbf{input:} \textcolor{gray}{\small\textit{node features $X$ of hypergraph $\mathcal{H}$ and ref. distribution q}}
      \State \textbf{output:} \textcolor{gray}{\small\textit{updated node features $\tilde{X}$}}
      \vspace{2mm}
      \Procedure{WHNN}{$X,\mathcal{H},q$} \\
        \vspace{2mm}
        \State $X_0 \gets X$
        \vspace{2mm}
        \State $\textcolor{gray}{\small\textit{\# Sample reference sets}}$
        \State $Q_v, Q_e \gets sample(q)$
        \vspace{2mm}
        \State $\textcolor{gray}{\small\textit{\# Extract node and edge neighbourhood}}$
        \State $\mathcal{N}_v, \mathcal{N}_e \gets \text{neighbourhoods}(\mathcal{H})$

        \vspace{2mm}  
        \State $\textcolor{gray}{\small\textit{\# Node to hyperedge}}$  
        \State $X \gets encoder(X)$ 
        
        \State $Z \gets Wasserstein(X, \mathcal{N}_v, Q_v)$ 
        \vspace{2mm}
        \State $\textcolor{gray}{\small\textit{\# Hyperedge to node}}$  
        \State $Z \gets encoder(Z)$ 
        \State $X \gets Wasserstein(Z, \mathcal{N}_e, Q_e)$

        \vspace{2mm}
        \State $\textcolor{gray}{\textit{\# Residual connection}}$  
        \State $\tilde{X} \gets \alpha X + (1-\alpha)X_0$
        
      \vspace{3mm}
      \State \textbf{return} $\tilde{X}$\
    \EndProcedure
    \footnotetext{$^*$ For simplicity in handling shapes, we assume encoders that are independent of the hyperedge.}
    \end{algorithmic}
    


\end{minipage}
\hfill
\begin{minipage}[t]{.49\textwidth}
    \captionof{algorithm}{Wasserstein aggregator}\label{algo2}
    \label{alg:Wass}
    \begin{algorithmic}[1]
      \State \textbf{input:} \textcolor{gray}{\textit{entity features $X$; list of neighbourhoods to aggregate $\mathcal{N}$; samples from reference distribution $Q$}}
      \State \textbf{output:} \textcolor{gray}{\textit{aggregated neighbourhoods $Z$}} 
      \vspace{2mm}
      \Procedure{Wasserstein}{$X, \mathcal{N}, Q$}
          \vspace{2mm}
          \State $\textcolor{gray}{\small\textit{\# Project entities into slices}}$ 
          \State $X \gets X\Theta$
          \State $\textcolor{gray}{\small\textit{\# Sort the samples from the reference distr.}}$ 
          \State $Q \gets \text{sort}(Q)$
          \vspace{2mm}
          \ForAll{$\textbf{ neighbourhoods } S \in \mathcal{N}$}
          \vspace{2mm}
            \State $\textcolor{gray}{\small\textit{\# Extract elements in the neighbourhood}}$
            \State $X_s \gets \{x_i \}_{i \in S}$
            \vspace{2mm}
            \State $\textcolor{gray}{\small\textit{\# If $|X_s|\neq|Q|$ interpolate $X_s$ to match size}}$
            \State $X_s' \gets \text{interpolate}(X_s)$
            \vspace{2mm}
            \State $\textcolor{gray}{\small\textit{\# Sort the elements of the neighbourhood.}}$ 
            \State $X_s' \gets \text{sort}(X_s')$
            \vspace{2mm}
            \State $\textcolor{gray}{\small\textit{\# Compute the dist that approx Wass dist}}$
            \State $Z_{s:} \gets Q - X_s'$
            
          \EndFor

          \vspace{2mm}
          \State $\textcolor{gray}{\small\textit{\# Combine the slices}}$
          \State $Z \gets ZW$
          \vspace{3mm}
          \State \textbf{return } $Z$\
          \vspace{3mm}
           \hrule  
      \EndProcedure
    \end{algorithmic}
\end{minipage}
}

\end{figure}

\paragraph{Hyperedges as probability distributions.}
    Unlike traditional hypergraph approaches that treat a hyperedge as a set of nodes, we model a hyperedge as a probability distribution, with its constituent nodes being samples drawn from that distribution. This way the hyperedges are not only characterized by the combination of its elements, but by the regions of the space where its elements are situated. The nodes became prototypes of the hyperedge behaviour. For example, a hyperedge in which nodes have similar representations ( homophilic behaviour) indicates a low-variance distribution while a hyperedge with diverse nodes suggests a more uniform distribution.


    
    
    
    Lets consider $p_i$ the probability distribution where the elements of the hyperedge $e_i$ are sampled from. In other words, we assume each node $v_j \in e_i$  is sampled as $\tilde{x}^i_j \in \mathbb{R}^d \sim p_i$. The goal is to obtain hyperedge embeddings that preserve the geometric information of this underlying distribution, such as spreading, shape etc. See Figure~\ref{fig:main} for a visual representation of this data structure. 
    
    Note that, by treating nodes as sampled from an underlying distribution, we make the assumption that other unobserved nodes drawn from the same distribution are also likely to belong to the same hyperedge. This probabilistic interpretation proved to be powerful for set representation learning~\cite{Wass_pooling} and our experiments demonstrates that hypergraph models can benefit from it as well.

\paragraph{Wasserstein aggregator.} Interpreting hypergraphs as a collection of probability distributions enable us to derive more powerful similarity metrics between hyperedges. As showed in the previous section, most of the current hypergraph architectures rely on mean pooling to create hyperedge embeddings from node representations. However, from a probabilistic perspective, averaging compares distributions only based on their means.  For complex data distributions, this approach fails to capture the full underlying geometry. While models relying on summation such as Deep Sets~\cite{DeepSets} where proved to be universal approximators, they heavily rely on the internal node encoder (an MLP) to map input features into a space where first-order statistics like the mean effectively approximate the distribution.  In the hypergraph setting, where multiple sets interact in complex ways, this is hard to achieve. 

This motivates us to adopt Sliced Wasserstein Pooling~\cite{Wass_pooling} to encode the hyperedge distributions. Concretely, for each hyperedge $e$, given the node embeddings of all the nodes in the hyperedge $\{~\tilde{x}_i^e \}_{i \in e}$, we are aggregating them using the Sliced Wasserstein Pooling described in Section~\ref{sec:Wass}, to obtain a vectorial hyperedge representation: $h_e = \mathcal{SWP}( \{ ~\tilde{x}_i^e \}_{i \in e} $). The algorithm works as follow:
\begin{enumerate}
    \item \textbf{Step 1}: Select a reference hyperedge distribution $q$ and sample $N$ points $\{ y_i\}_{i=1}^N \sim q$. Choose a set of directions $\{ \theta_l \}_{l=1}^{L}$ with $\theta_l \in \mathbb{R}^{d \times 1}$ used as projection slices in the pooling process. Note that, in order to obtain comparable embeddings across the entire hypergraph, we share the same reference distribution  and the same set of slices for all hyperedges.

    \item \textbf{Step 2}: Project each node representation $\tilde{x}_i^e$ into each slice $\theta_l$ as follow: $z_i^{e, \theta_l} = (\tilde{x}_i^e)^T \theta_l \in \mathbb{R}$. Since the algorithm requires the same number of sampled nodes from both the hyperedge distribution and the reference, when the cardinality of the hyperedge  $|e| \neq N$, we increase/decrease the number of nodes in $e$ using linear interpolation. $z_i^{e, \theta_l} \gets \text{interp}(z_i^{e, \theta_l}, N)$

    \item \textbf{Step 3}: For each hyperedge, for each slice, compute the distance between the node representations and the reference points. $h_e^{\theta_l} = ||z_{\pi(i)}^{e,\theta_l} - y_{\tilde{\pi}(i)}||$, where $z^{e,\theta_l}_\pi$ and $y_{\tilde{\pi}}$ represent the vectors in sorted order. The final hyperedge embedding is obtain as a weighted mean of these embeddings: $h_e = \sum_{l=1}^{L}(w_l h_e^{\theta_l})$, where $w_l$ are learnable scalars combining the slices.
\end{enumerate}
Intuitively, each hyperedge is represented by a vector which measure how difficult it is to transform the \textit{hyperedge distribution} into the reference distribution~\footnote{As defined above, by hyperedge distribution we denote the distribution of nodes in the hyperedge.}. Following the theoretical properties of Sliced Wasserstein Pooling~\cite{Wass_pooling}, the Euclidean distance between two hyperedge representations measures the cost of transforming one hyperedge distribution into another. Hyperedges that are similar in shape or spreading should be closer in this space compared to hyperedges that have completely different  distributions. The algorithm is also described in Algorithm~\ref{alg:Wass}. The directions $\theta_l$ and the reference distribution can be either fixed or learnable.

\paragraph{Edge to node step.}
For simplicity, we only described in details the first stage of the framework which sends messages from nodes to hyperedges. The second stage of the framework which create node representation by aggregating the information from neighbouring hyperedges is done in a similar way, only with different parameters. In conclusion, we not only capture structural relationship between hyperedges, but also structural relationship between nodes' neighbourhood.  


\begin{figure}[t]
    \centering
     \includegraphics[width=0.9\textwidth]{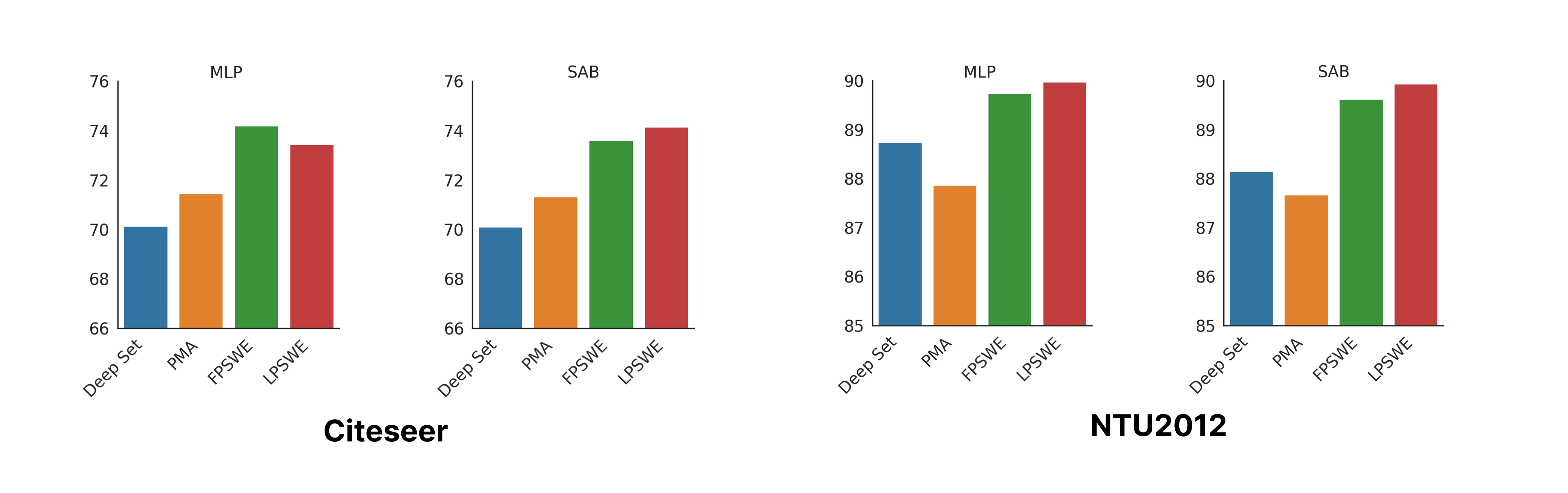}
     \caption{\textbf{Ablation study on the importance of Wasserstein aggregator for hypergraph representation learning} on Citeseer and NTU2012 datasets. We are testing two versions of the Sliced Wasserstein Pooling: with fixed (FPSWE) or learnable (LPSWE) reference distribution. Regardless of the encoder used to project the nodes and hyperedges, the Wasserstein aggregators outperform both the Deep Sets and PMA commonly used inside hypergraph models.}
     \label{fig:visual_ablation}
     \vspace{-3mm}
\end{figure}

\section{Experiments}
\label{experiments}

Our main goal is to understand to what extent Wasserstein aggregation is beneficial for hypergraph neural networks. Additionally, we investigate how the choice of node encoder—whether edge-dependent or edge-independent—affects overall performance. Finally, we compare our model against a range of strong baseline methods from the existing literature.

\paragraph{Datasets.} We evaluate our model on the node-classification task. 
We select seven real-world datasets that vary in domain and scale. These include Cora, Citeseer, Cora-CA, DBLP-CA~\cite{HyperGCN},  ModelNet40~\cite{modelnet}, NTU2012~\cite{ntu2012} and 20News~\cite{20News}. Among the datasets that are usually used for benchmarking hypergraph models~\cite{HyperGCN}, we omitted Pubmed due to the high percentage of isolated nodes ($80.5\%$) which makes the relational processing unnecessary. Senate and House are two other datasets used to test hypergraphs in the heterophilic regime. However, they do not provide node features which are a key component for our geometric and probabilistic interpretation.

For a fair comparison with the other methods, we follow the training procedures employed by~\cite{wang2022equivariant}. We randomly split the data into $50\%$ training samples, $25\%$ validation samples and $25\%$ test samples.


\paragraph{Importance of Wasserstein aggregator.} Our main contribution consists of adopting Sliced Wasserstein Pooling as a powerful aggregator inside the hypergraph networks. As described in the previous section, while most of the existing methods used variations of the sum pooling to aggregate the information from each node and each hyperedge neighbourhoods, our Wasserstein aggregator presents a more in-depth understanding of the neighbourhood distribution, being capable of capturing subtle differences such as the difference in shape or spread. 

To understand to what extent this is contributing to a better hypergraph representation for real-world scenarios, we are designing an ablation study in which we keep the underlying architecture fixed and only modify the aggregator used in both the nodes-to-hyperedges and hyperdges-to-nodes stages. Concretely we are using as aggeregators either Deep Set module (as used by AllDeepSet and ED-HNN models) or the PMA module (as used by AllSetTransformer model). For our Wasserstein aggregator, we are experimenting with both a fixed-reference distribution (a model denoted as FPSWE) or with learnable reference distribution (a model denoted as LPSWE). For a robust evaluation we are comparing this aggregators using both the edge-independent encoder (MLP) and  the edge-dependent encoder (SAB). The results on Citeseer and NTU2012 datasets are reported in Figure~\ref{fig:visual_ablation}. 

Regardless of the encoder and the dataset we are testing on, both Wasserstein aggregators are consistently outperforming both the Deep Sets and the PMA aggregators by a significant margin. A learnable reference seems to be beneficial, however the improvement is generally marginal. 
Additional experiments on other datasets show a similar trend and are provided in the Supplementary Material.

\paragraph{Importance of edge-dependent encoder.} The node and hyperedge encoder transforms features into a space where their distribution within each hyperedge captures meaningful information about the group. As stated in the model description, we equipped our model with two types of encoders. An edge-independent module represented by an MLP, and an edge-dependent encoder represented by a self-attention block (SAB). While the MLP is processing information independently for each node/hyperedge, SAB is capturing pairwise interactions between nodes/hyperedges sharing a neighbourhood.  
The results in Figure~\ref{fig:visual_ablation} and Table~\ref{tab:sota} show similar results among the encoder, with the edge-dependent one being slightly more powerful. However, this comes with the cost of a more expensive model, as the edge-dependent encoder requires more memory to store the representation for all incident pairs (node, hyperedge). To alleviate that on the larger datasets (20News and DBLP), we replace the SAB block with the ISAB low-rank approximation introduced by ~\cite{set_transformer}.

\begin{table}[t]
  \caption{\textbf{Performance on a collection of hypergraph datasets.} 
  Our model using SWP as a node and hyperedge aggregator shows superior results, proving the advantage of moving beyond the standard sum pooling employed by most of the existing works. We test our model in both its variants: with edge-independent (MLP) and edge-dependent encoder (SAB). Both options are exhibiting competitive performance. We mark the \textbf{first}, \underline{second} and \textit{third} best performing models for each dataset.\\}
  
  \label{tab:sota}
  \centering
  \resizebox{\columnwidth}{!}{%
  \begin{tabular}{cccccccccc}
    \toprule
     Name      & Cora & Citeseer & Cora\_{CA} & DBLP\_{CA} & ModelNet40 & NTU2012 & 20News  \\
    \midrule
     HCHA & 79.14 $\pm$ 1.02 & 72.42 $\pm$ 1.42 & 82.55 $\pm$ 0.97 & 90.92 $\pm$ 0.22  & 94.48 $\pm$ 0.28 & 87.48 $\pm$ 1.87 & 80.33 $\pm$ 0.80 &   \\
     HNHN & 76.36 $\pm$ 1.92 & 72.64 $\pm$ 1.57  &  77.19 $\pm$ 1.49 & 86.78 $\pm$ 0.29 & 97.84 $\pm$ 0.25 & 89.11 $\pm$ 1.44 & 81.35 $\pm$ 0.61 & \\
     HyperGCN & 78.45 $\pm$ 1.26 & 71.28 $\pm$ 0.82 & 79.48 $\pm$ 2.08 & 89.38 $\pm$ 0.25 & 75.89 $\pm$ 5.26 & 56.36 $\pm$ 4.86& 81.05 $\pm$ 0.59 & \\
     HyperGNN & 79.39 $\pm$ 1.36 & 72.45 $\pm$ 1.16  & 82.64 $\pm$ 1.65 & 91.03 $\pm$ 0.20 & 95.44 $\pm$ 0.33 & 87.72 $\pm$ 1.35 & 80.33 $\pm$ 0.42&  \\
     AllDeepSets & 76.88 $\pm$ 1.80 & 70.83 $\pm$ 1.63 & 81.97 $\pm$ 1.50 & 91.27 $\pm$ 0.27 & 96.98 $\pm$ 0.26 & 88.09 $\pm$ 1.52 & 81.06 $\pm$ 0.54 &\\
     AllSetTransformers & 78.58 $\pm$ 1.47 & 73.08 $\pm$ 1.20 & 83.63 $\pm$ 1.47 & 91.53 $\pm$ 0.23 & \textit{98.20 $\pm$ 0.20} & 88.69 $\pm$ 1.24 & \textit{81.38 $\pm$ 0.58} & \\
     UniGCNII & 78.81 $\pm$ 1.05 & 73.05 $\pm$ 2.21 & 83.60 $\pm$ 1.14 &  91.69 $\pm$ 0.19 & 98.07 $\pm$ 0.23 & 89.30 $\pm$ 1.33 & 81.12 $\pm$ 0.67 &  \\
      ED-HNN & \underline{80.31 $\pm$ 1.35} & \textit{73.70 $\pm$ 1.38} & \textit{83.97 $\pm$ 1.55} & \underline{91.90 $\pm$ 0.19} & 97.75 $\pm$ 0.17 & \textit{89.48 $\pm$ 1.87} & $81.36 \pm 0.55$ &  \\
     \midrule
     WHNN\_MLP & \textit{79.84 $\pm$ 1.56}  & \underline{74.79 $\pm$ 1.19} & \underline{84.12 $\pm$ 1.94} & \textit{91.73 $\pm$ 0.24} &  \underline{98.47 $\pm$ 0.19} & \textbf{\text{90.87 $\pm$ 1.59}} & \textbf{\text{81.83 $\pm$ 0.68}} \\
     WHNN\_(I)SAB & \textbf{\text{80.72 $\pm$ 1.96}}  & \textbf{\text{74.92 $\pm$ 1.60}} & \textbf{\text{84.62 $\pm$ 1.77}} & \textbf{\text{91.99 $\pm$ 0.33}} & \textbf{\text{98.54 $\pm$ 0.21}} & \underline{90.68 $\pm$ 1.68} & \underline{81.42 $\pm$ 0.60} \\
    
    \bottomrule
  \end{tabular}%
  }
\end{table}
\noindent\textbf{Comparison with baselines.} In Table~\ref{tab:sota} we are comparing against a series of hypergraph networks from the literature. With respect to aggregation strategies, HNHN~\cite{hnhn}, HyperGNN~\cite{HyperNN}, AllDeepSets~\cite{allset}, UniGCNII~\cite{UniGNN} and ED-HNN~\cite{wang2022equivariant} use variations of Deep Sets to aggregate the information, HyperGCN~\cite{HyperGCN} uses a max aggregator, while HCHA~\cite{HCHA} and AllSetTransformer~\cite{allset} are using an attention-based weighted summation. Regardless of the encoder used, our model consistently obtain top results, outperforming the other methods on all datasets. This demonstrates the advantages of using Wasserstein aggregators for higher-order processing. Note that, while we integrated this aggregator into a standard instantiation of the two-stage framework, many existing models from the literature can be adopted to take advantage of this type of geometric-inspired aggregation.

\noindent\textbf{Implementation details.} In all experiments, we train our models using Adam optimizer for $500$ epochs, on a single GPU NVIDIA Quadro RTX 8000 with 48GB of memory. For comparing against other models in the literature, each model is trained $10$ times with different random splits and different initialization. For each experiment we report average accuracy along with the standard deviation. The results represent the best performance obtained by each architecture using hyper-parameter optimisation with random search. For the ablation study the results are averaged across $5$ runs and the architecture is fixed to ensure a fair comparison. For all experiments we use a number of Wasserstein slices equal to the hidden dimension and we experiment with both learning the reference set or not. Details on all the model choices and hyper-parameters can be found in the Supplementary Material.

These experimental results show that aggregating node and hyperedge neighborhoods using Sliced Wasserstein Pooling is highly effective for hypergraph processing, the Wasserstein aggregator consistently outperforming standard methods like Deep Sets and PMA.

\section{Conclusion}
In this work we introduce Wasserstein Hypergraph Neural Networks (WHNN), a model for processing hypergraph structures. The model relies on Sliced Wasserstein Pooling to aggregate the nodes into hyperedge representations and vice versa. This design choice inspired by optimal transport literature enable us to capture more information about the internal structure of the neighbourhoods, preserving more geometric relation between elements. The experimental results on various datasets demonstrates that this Wasserstein aggregator is effective for modeling higher-order interactions, outperforming traditional aggregators, making WHNN a promising tool for hypergraph representation learning.

\paragraph{Acknowledgment.} The authors would like to thank Daniel McFadyen and Alex Norcliffe for fruitful discussions and constructive suggestions during the development of the paper. I.D. additionally acknowledges the support provided by Robinson College during this period. 


\bibliographystyle{unsrt}
\bibliography{main}

\newpage
\appendix

\section*{\centering\LARGE\bf Appendix: Wasserstein Hypergraph Neural Network}\vspace{12mm}

This appendix contains details related to our model, including potential limitations and future work, additional datasets for the ablation experiments, details on the hyperparameters used in our experiments and derivation of the computational complexity. The appendix is structured as follow:

\begin{itemize}
    \item \textbf{Section A} highlights a series of potential limitations that can be address to improve the current work, together with a discussion on potential future work.
    \item \textbf{Section B} presents additional experiments used as ablation for our model.
    \item \textbf{Section C} presents the list of hyperparameters used in our experiments.
     \item \textbf{Section D} derives the computational complexity of our model.

\end{itemize}

\section{Limitations and Future work}

As discussed in the main paper, we treat the neighborhood of each node as a sample from an underlying probability distribution. This approach assumes that any additional nodes drawn from this distribution should belong to the same neighborhood as the observed ones. This aligns with the intuition that elements within a group should share common characteristics. While the datasets we used support this assumption, there may be real-world scenarios where it does not hold. Our model relies solely on the node encoder to project features into a space where the assumption is approximately valid. 

Moreover, due to this continuous view of the neighbourhood (as a distribution of probability) together with the interpolation step , the current model may lose information about the exact cardinality of the neighborhoods. In situations where neighborhood size is important, we recommend encoding it as an explicit feature. However, we mention that this is an issue we share with the mean-based pooling algorithms.  

The main goal of this paper is to highlight the benefits of using geometrically-inspired poolings for aggregating neighbourhood information in hypergraphs. While we focused entirely on hypergraphs, similar idea can be apply on graph neural networks to aggregate messages coming from each node's neighbourhood. As a future work, it would be interesting to see to what extent GNNs can benefit from Wasserstein aggregators.

Moreover, while the proposed model integrate the Wasserstein aggregator into a standard two-stage pipeline, several other architectures such as ED-HNN that uses summation as an aggregator might benefit from adopting it. We are leaving this investigation as future work.

\section{Additional experiments}

Due to space constraints, in the main paper we only included ablation studies on Citeseer and NTU datasets. Here we report additional results for Cora\_CA and ModelNet40 datasets. 

For each experiment, we kept the architecture fixed and modify the aggregator used in the two stages to be either Deep Set, PMA, and the learnable (LPSWE) or fixed (FPSWE) Wasserstein aggregator. The results are similar across the datasets, with Wasserstein Pooling proving to be beneficial compared to Deep Sets and PMA. In terms of encoder type, we noticed that, in some cases, for a fixed architecture, SAB tends to model the distribution better than MLPs.

\begin{figure}[h]
    \centering
     \includegraphics[width=0.95\textwidth]{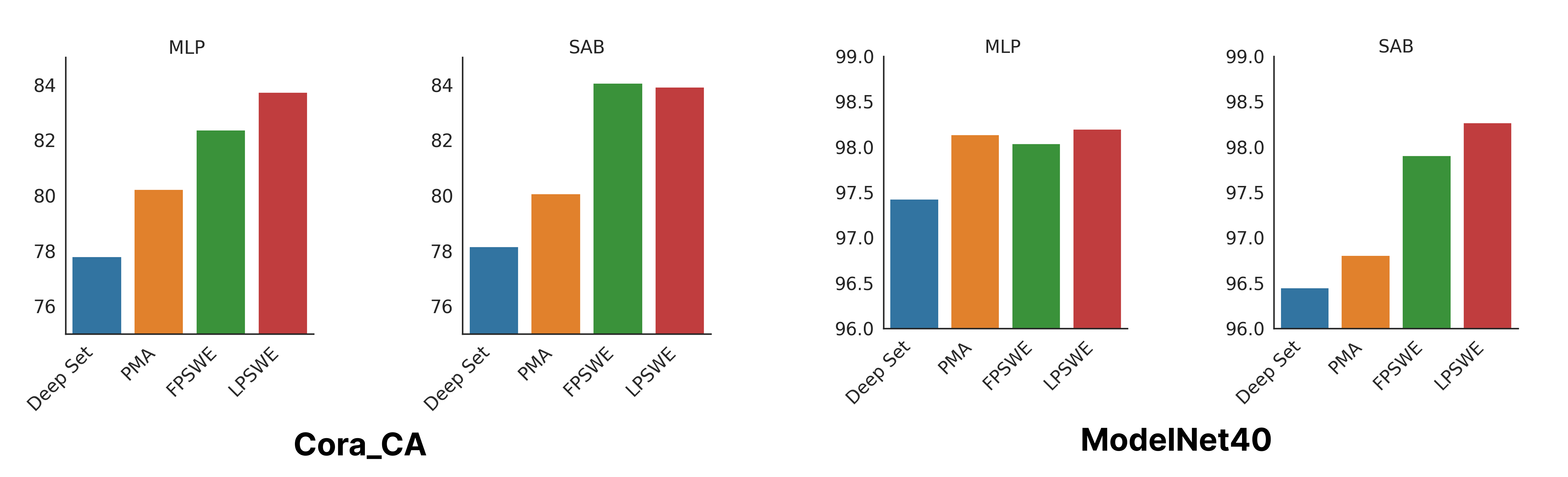}
     \caption{\textbf{Additional results for the ablation study on the importance of Wasserstein aggregator for hypergraph representation learning} Cora\_CA and ModelNet datasets. FPSWE denotes the Wasserstein aggregator with fixed reference while LPSWE denotes the Wasserstein aggergator with learnable reference distribution. Regardless of the encoder used to project the nodes and hyperedges, the Wasserstein aggregators outperform both the Deep Sets and PMA commonly used inside hypergraph models.}
     \label{fig:visual_ablation2}
     \vspace{-3mm}
\end{figure}

\section{Implementation details}

The results reported in Table 2 of the main paper are obtained using random hyperparameter tuning. We report here the range of parameters that we searched for. Table~\ref{tab:hyperparam_MLP} and Table~\ref{tab:hyperparam_SAB} contains the best hyperparameter configuration for the WHNN\_MLP model and WHNN\_SAB. We depict in bold the parameters specific to the Wasserstein aggregator, in italic the parameters specific to the SAB encoder, while the rest of them are the standard parameters used in the two-stage hypergraph models. In our experiment we search for the following hyperparameters:

\begin{itemize}
    \item num\_ref: number of elements sampled from the reference distribution $\{ 5, 10, 25, 50 \}$
    \item learnable\_W: choose between learning or not the reference distribution $\{ \text{True}, \text{False} \}$
    \item heads: number of heads used by the SAB block $\{ 1, 2, 4 \}$
    \item MLP\_layers: number of layers in all MLPs used $\{ 0, 1, 2 \}$
    \item MLP\_hid: number of hidden units in all MLPs used. This is also the number of slices used by Wasserstein aggregator. $\{ 128, 256, 512 \}$
    \item MLP2\_layers: using or not an additional linear projection after the residual connection of each stage $\{ 0, 1 \}$
    \item Cls\_layers: number of layers in the final classifier MLP $\{ 1, 2 \}$
    \item Cls\_hid: number of hidden units in the final classifier MLP $\{ 96, 128, 256 \}$
    \item self\_loops: using or not self loops $\{ \text{True}, \text{False} \}$
    \item dropout: dropout used inside the model $\{ 0.5, 0.6, 0.7 \}$
    \item in\_dropout: dropout used in the begining of the model  $\{ 0.2, 0.5, 0.6, 0.7 \}$
    \item fixed hyperparameters: All models use 1 layer of WHNN, LayerNorm normalisation, the residual coefficient $\alpha$ fixed to $0.5$ and they are trained for $500$ epochs  with a learning rate of $0.001$.   
\end{itemize}

\begin{table}[h!]
\caption{The best configuration of hyperparameters used by our model WHNN\_MLP on all tested datasets. We mark with bold the parameters that are specific to the Wasserstein aggregator.}
\vspace{2mm}
\label{tab:hyperparam_MLP}
  \centering
  \resizebox{\columnwidth}{!}{%
\begin{tabular}{lccccccc}
\toprule
\text{Parameter} & \text{Cora} & \text{Citeseer} & \text{Cora\_CA} & \text{DBLP\_CA} & \text{ModelNet40} & \text{NTU2012} & \text{20News} \\
\midrule
\textbf{\text{num\_ref}}       &  25&  10&  25&  5&  50&  25&  25\\
\textbf{\text{learnable\_W}}   &  True&  False&  True&  True&  False&  False&  False\\
\text{MLP\_layers}      &  1&  2&  2&  2&  1&  1&  0\\
\text{MLP2\_layers}     &  0&  0&  1&  0&  0&  1&  0\\
\text{MLP\_hid}         &  128&  256&  256&  512&  256&  512&  512\\
\text{Cls\_layers}      &  1&  1&  1&  2&  2&  2&  2\\
\text{Cls\_hid}         &  256&  128&  96&  96&  96&  96&  96\\
\text{self\_loops}      &  True&  True&  True&  True&  True&  False&  False\\
\text{dropout}          &  0.7&  0.5&  0.6&  0.7&  0.5&  0.5&  0.5\\
\text{in\_dropout}      &  0.7&  0.5&  0.6&  0.7&  0.2&  0.2&  0.2\\
\bottomrule
\end{tabular}
}

\end{table}

\begin{table}[h!]
\caption{The best configuration of hyperparameters used by our model WHNN\_SAB on all tested datasets. We mark with bold the parameters that are specific to the Wasserstein aggregator and with italic the parameters that are specific to the SAB encoder.}
\vspace{2mm}
\label{tab:hyperparam_SAB}
  \centering
  \resizebox{\columnwidth}{!}{%
\begin{tabular}{lccccccc}
\toprule
\text{Parameter} & \text{Cora} & \text{Citeseer} & \text{Cora\_CA} & \text{DBLP\_CA} & \text{ModelNet40} & \text{NTU2012} & \text{20News} \\
\midrule
\textbf{\text{num\_ref}}       &  10&  5&  50&  5&  25&  25&  5\\
\textbf{\text{learnable\_W}}   &  True&  False&  False&  False&  False&  False&  True\\
\textit{heads}        &  2&  4&  1&  4&  1&  2&  2\\
\text{MLP\_layers}      &  2&  2&  2&  1&  1&  2&  2\\
\text{MLP2\_layers}     &  0&  0&  1&  1&  0&  0&  0\\
\text{MLP\_hid}         &  128&  256&  128&  256&  256&  512&  512\\
\text{Cls\_layers}      &  1&  1&  1&  2&  2&  2&  2\\
\text{Cls\_hid}         &  128&  256&  128&  96&  96&  96&  96\\
\text{self\_loops}      &  True&  False&  True&  True&  True&  True&  False\\
\text{dropout}          &  0.7&  0.7&  0.5&  0.7&  0.5&  0.5&  0.5\\
\text{in\_dropout}      &  0.7&  0.7&  0.5&  0.7&  0.2&  0.2&  0.2\\
\bottomrule
\end{tabular}
}

\end{table}

\newpage
\section{Computational complexity}
We derive the computational complexity for both versions of our Wasserstein Hypergraph Neural Network: using the edge-independent encoder (WHNN\_MPN) and using the edge-dependent encoder (WHNN\_SAB). We present the complexity for a hypergraph with $N$ nodes, $M$ hyperedges, $K_e$ the maximum cardinality of a hyperedge, $K_v$ the maximum number of hyperedges a node is part of and $R$ the number of reference points sampled.

For node encoders, the MLP encoder has a complexity of $O(N)$ while the SAB encoder has complexity $O(M \times K^2)$ due to the pairwise exchange of messages ($K^2$) inside each hyperedge ($M$).

For the Wasserstein aggregator, we derive the complexity both for the {node-to-hyperedge} and {hyperedge-to-node} stages. For {node-to-hyperedge} the complexity for interpolation is $O(M \times (R \log K_e))$ and the complexity for sorting each neighbourhood is $O(M \times (R \log R))$. Symmetrically, for {hyperedge-to-node} the complexity for interpolation is $O(N \times (R \log K_e))$ and the complexity for sorting each neighbourhood is $O(N \times (R \log R)$. The overall complexity becomes $O(M \times (R \log K_e) + M \times (R \log R)) + N \times (R \log K_e) + N \times (R \log R))$. In our experiments R is maximum $50$.

For comparison, the complexity of a Deep Set pooling is $O(M \times K_e + N \times K_v)$

\end{document}